\documentclass[11pt,twoside,a4paper]{article}

\usepackage{hyperref}

\newcommand{\longurl}[1]{%
    {\expandafter\def\expandafter\UrlBreaks\expandafter{\UrlBreaks\UrlOrds%
        \do\/\do\a\do\b\do\c\do\d\do\e\do\f%
        \do\g\do\h\do\i\do\j\do\k\do\l\do\m%
        \do\n\do\o\do\p\do\q\do\r\do\s\do\t%
        \do\u\do\v\do\w\do\x\do\y\do\z%
        \do\A\do\B\do\C\do\D\do\E\do\F\do\G%
        \do\H\do\I\do\J\do\K\do\L\do\M\do\N%
        \do\O\do\P\do\Q\do\R\do\S\do\T\do\U%
        \do\V\do\W\do\X\do\Y\do\Z}%
    \url{#1}}%
}

\usepackage{amsmath}
\usepackage{graphicx}
\usepackage{subcaption}
\usepackage{listings}
\usepackage{xcolor}
\usepackage{comment}
\usepackage{adjustbox}

\definecolor{gray}{rgb}{0.4,0.4,0.4}
\definecolor{darkblue}{rgb}{0.0,0.0,0.6}
\definecolor{cyan}{rgb}{0.0,0.6,0.6}
\lstdefinelanguage{XML}
{
  basicstyle=\ttfamily\scriptsize,
  morestring=[b]",
  morestring=[s]{>}{<},
  morecomment=[s]{<?}{?>},
  stringstyle=\color{black},
  identifierstyle=\color{darkblue},
  keywordstyle=\color{cyan},
  morekeywords={xmlns,aas,type}
}

\lstdefinestyle{turtle}{
  basicstyle=\ttfamily\scriptsize,
  breakatwhitespace=false,         
  breaklines=true,                 
  captionpos=b,                    
  keepspaces=true,                 
  numbers=left,                    
  numbersep=5pt,                  
  showspaces=false,                
  showstringspaces=false,
  showtabs=false,                  
  tabsize=2,
  keywordstyle=\color{cyan},
  morekeywords={rdf,rr,rml,rdfs,rami,xsd}
}

\begin{document}

\title{The Semantic Asset Administration Shell}
%
%
\author{Sebastian R. Bader and 
Maria Maleshkova\\
Fraunhofer Institute IAIS, \\Schloss Birlinghoven, 53757 Sankt Augustin, Germany\\
University of Bonn, Endenicher Allee 19a, 53115 Bonn, Germany}

\date{}
\maketitle              
\begin{abstract}
The disruptive potential of the upcoming digital transformations for the industrial manufacturing domain have led to several reference frameworks and numerous standardization approaches. On the other hand, the Semantic Web community has made significant contributions in the field, for instance on data and service description, integration of heterogeneous sources and devices, and AI techniques in distributed systems. These two streams of work are, however, mostly unrelated and only briefly regard each others requirements, practices and terminology. 
We contribute to closing this gap by providing the Semantic Asset Administration Shell, an RDF-based representation of the Industrie 4.0 Component. We provide an ontology for the latest data model specification, created a RML mapping, supply resources to validate the RDF entities and introduce basic reasoning on the Asset Administration Shell data model. Furthermore, we discuss the different assumptions and presentation patterns, and analyze the implications of a semantic representation on the original data. We evaluate the thereby created overheads, and conclude that the semantic lifting is manageable, also for restricted or embedded devices, and therefore meets the needs of Industrie 4.0 scenarios. 

\end{abstract}
%
%

%
%
\section{Introduction}

Even though the various digital developments and internet-based technologies have attracted great attention in the manufacturing industry, a common understanding of the resulting requirements and implications has not been reached. The number of different terms, which are being used in this context, reflects this challenge -- Internet of Things (IoT), Industrial Internet, Cyber-physical Systems, and more specifically, Digital Twins, Smart Components~\cite{keppmann2016smart}, Virtual Representations\cite{bader2018virtual}, Smart Services~\cite{maleshkova2016sws} and many more have slightly overlapping scopes but still depict different applications and features. Still, the primary target is always the effective integration and interoperability of industrial devices, services and data sources. Therefore, the actual implementations require clear specifications of the used data formats, interfaces, and semantic meaning of the referenced objects and attributes.

IoT data is currently mainly exchanged in either JSON or XML. These commonly used data formats ease the serialization and parsing by providing specifications for the syntactic structure of the data objects. Additional information on the meaning of keys/values is usually specified in customized data models and schemata. 
The latest specification of the Plattform Industrie 4.0 Asset Administration Shell (AAS) also follows this convention~\cite{aasindetail}. The AAS is promoted as the digital twin for the German Plattform Industrie 4.0 and encompasses the interpretation of the digital representation of any production-related asset. As such, materials and products, devices and machines but also software and digital services have a respective digital version.

While the predefined structure and the usage of specific keys reduce the heterogeneity inherent in the data exchange processes of current industrial scenarios,  all real-world scenarios still require a thorough understanding of the specific terms and values (e.g., dispatching processes for predictive maintenance~\cite{bader2017supporting}). Therefore they are dependent on extensive manual work and understanding of the extended AAS model, followed by a time consuming data mapping.
A semantic formalization of entities and data objects has several advantages in this context. The mature Semantic Web technology stack around RDF enables clear references to classes, properties and instances in the form of URIs, beyond the scope of single AAS objects but also across applications, domains, and organizations. The defined meaning of the used entities further allows its combination with predefined logical axioms, which allow the automatic derivation of new knowledge.                         

We contribute to the state of the art by presenting a mapping from the latest AAS data model to RDF. Thus we provide a data model as an openly accessible ontology and create SHACL shapes for all classes to enable schema validation. We outline the various pitfalls, especially the different patterns to identify, and refer to encoded entities and to links to remote resources. Based on the inherent Web nature of RDF, we show how the transformation to the semantic data model decreases the amount of required storage space.
Furthermore, we present patterns to directly insert the RDF translation into the original XML and JSON files and discuss their implications. Relying on the RDF/XML and JSON-LD serializations, we are able to merge the predefined data structure with the semantically defined data. We show that the provided extension points in the form of submodel elements are suitable for this task and that the output AAS files are still processable by existing software, therefore the risk of compatibility issues is manageable.


The applicability of the presented approach is evaluated by determining the necessary overhead in terms of both storage and computation effort, and by a detailed discussion of the restrictions of the RDF version. We show that some semantic constructs are more efficient than the originally specified ones, whereas others are not directly compatible with the data structure of RDF and some are even not expressible at all. 

In this context the paper makes the following contributions:  
(1) an RDF data model of the Semantic Asset Administration Shell $SAAS$, 
(2) a mapping from XML Asset Administration Shell representations to $SAAS$, 
(3) a set of preliminary reasoning axioms in order to explicitly derive implicitly encoded information from the data model, and 
(4) a validation model for this data model, encoded through SHACL shapes.

The remainder of this paper is organized as follows. Section 2 contains an overview on similar efforts in the field. Section 3 introduces a formalization of the regarded domain followed by the presentation of the RAMI ontology and an RML mapping in Section 5. Section 6 briefly examines several axioms for automated reasoning on top of the SAAS, while Section 7 illustrates the provided SHACL Shapes for schema validation. We use several use cases (Section 8) to evaluate our approach (Section 9). Finally, we conclude with a discussion on the potential of the SAAS and outline further research gaps.

%
%
\section{Related Work}

In this section, we discuss three areas of related work -- the data model of the Asset Administration Shell, the existing mappings towards a semantic representation and related mappings of Industrie 4.0 data models to RDF.

Barnstedt et al. define the data model of the Asset Administration Shell~\cite{aasindetail}, the form of identifiers, access rights and roles, as well as XML and JSON serializations and their transport. The textual documentation of the model is enhanced with XML and JSON schemata. The model defines a basic set of keys and properties, and outlines defined points for custom vocabularies and terminologies. Part 2 of specification will further determine the APIs and interaction functions of the Asset Administration Shell, and how operations can be provided and described for the Industrie 4.0.

\vspace{-1em}
\begin{figure}[h]
        \centering
        \includegraphics[width=1.0\linewidth]{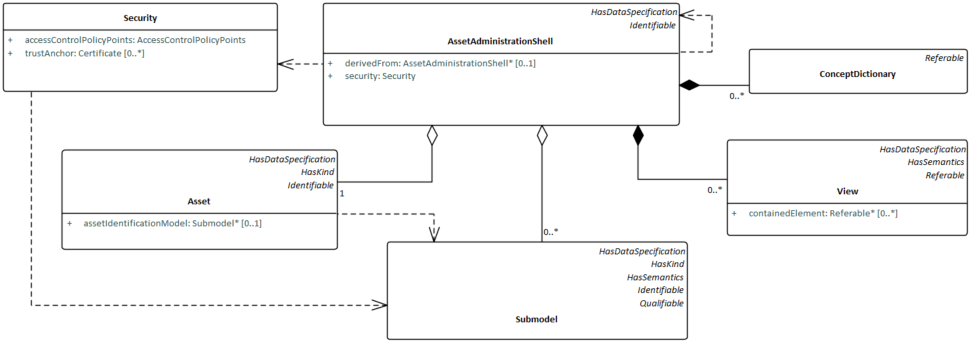}
        \caption{Sections of the Asset Administration Shell Data Model according to~\cite{aasindetail} (page 44). } 
        \label{fig:AasDataModel}
\end{figure}

Grangel-Gonz\'alez provide a first RDF data model for the Administration Asset Shell and the respective technical standards as published by ISO, IECC, and DIN~\cite{grangel2016towards}. They further extended the work in~\cite{grangel2016rdf} with a formalized model of the Reference Architecture for Industrie 4.0 (RAMI4.0) and entities for units of measurements and provenance, and show a prototypical mapping using R2RML. However, the mapping itself was not generally applicable to other Asset Shells as a common data model was not specified at this time.

Tantik and Anderl~\cite{tantik_integrated_2017} present an analysis how recommendations of the World Wide Web Consortium (W3C) fit to the guidelines of the Plattform Industrie 4.0. They outline various suggestions how standardized Web technologies can be integrated into Asset Shells. The authors present best practices and integration methods through a sample implementation scenario but do not discuss the implications on the data model itself.

Mappings of relational or otherwise formatted data to RDF are possible with the RDB to RDF Mapping Language R2RML~\cite{das2012r2rml} or the broader applicable RDF Mapping Language RML~\cite{dimou2014rml}, which also enables mappings from JSON, XML or CSV to RDF. The desired transformations are also formulated in RDF by defining the output graph structure by so-called Maps and URI templates. While R2RML strictly relies on tables, and uses column names as resource and attribute identifiers of row-based data objects, RML also transforms JSON and XML data by identifying objects according to their keys. Even though some tools have been introduced in order to support the creation of mappings for both approaches, the possibility to collaboratively work on mappings was not part of the design requirements and is still missing.

Katie et al.~\cite{katti_sa-opc-ua_2018} show by integrating the machine-to-machine communication protocol OPC-UA for servers and clients how semantic descriptions, in particular SAWSDL annotations, bridge the gap between the heterogeneous devices of the shop floor. The use of uniquely identified semantic descriptions supports the automatic orchestration of decoupled Cyber-physical Systems. However, only the specific input and output requirements of the OPC-UA methods are described. Neither the data objects nor the OPC-UA general information model is reflected.

Dietrich et al. examine the semantic characteristics of the Asset Administration Shell in~\cite{diedrich_semantic_2017}. They outline the identification of attributes and properties through cross-industry standards, mainly IEC 61360 and eCl@ss. In addition, they discuss mappings to AutomationML and OPC-UA. However, Dietrich et al. do not recognize the concepts of the Semantic Web and therefore do not show how to integrate the Administration Shell with its technology stack.

Currently, to the best of our knowledge, there is no RDF representation of the officially released data model of the Asset Administration Shell. This is necessary in order to build a bridge between the the latest approaches of data provisioning models in the manufacturing domain and the rich and mature data integration and formalization capabilities of the Semantic Web. As such, an RDF data model has the potential to ease the information exchange but also provides the capabilities to introduce logical reasoning to the Asset Administration Shell.

%
%
\section{Methodology}
The data model for the Industrie 4.0 component aims to provide high coverage of the different modeling variants. RDF on the other hand has specific conditions how data is presented (triple-based structure, URI as identifier). In order to structure the contribution of this paper, the parts of the respective data models are defined as follows:

\textbf{\textit{AAS}} captures the information about the Administration Asset Shell itself. In this regard, \textit{AAS} is the digital representation or Digital Twin of the Asset. Information from \textit{AAS}, therefore, refers to the information object or document and only indirectly to the original asset. Examples are the creation date of the digital representation, manuals, or how the AAS was generated or modified. It is important to note that the same reference is used to denote both the Administration Asset Shell itself and the set of information contained by it.

\textbf{\textit{A}} captures the information about the actual asset. The asset can be anything of interest in the context of a digital production setting. Even though assets are usually embedded devices or internet-capable components, any physical object, such as materials, production goods or machines, can be seen as an asset too. In addition, assets also include software components and any digital service or intangible thing, which is necessary to model a manufacturing use case. 

\textbf{\textit{S}} denotes the submodel of the asset shell. Submodels partition the provided information and categorize facts according to their usage, for instance as part of a documentation submodel or a submodel for quality testing. Submodels are further separated into SubmodelElements, which are either themselves collections of SubmodelElements or the final bearer of key-value-encoded facts. As any combination of different submodels can be included in the Asset Administration Shell, the set $S^k$ represents the superset, including all possible submodels.

\textbf{\textit{I}} is the set of identifiers for data objects. Specifically $I = I_{glob} \cup I_{loc} $ where $I_{glob}$ contains all globally valid identifiers, while the elements of $I_{loc}$ are only valid in their context, in particular inside the AAS, which uses them. 

The concept descriptions denoted with \textbf{\textit{CD}} may provide further definitions about the used concepts, mainly attributes and data types. While concept descriptions are optional components of an AAS, they give the ability to place necessary explanations especially for entities with local identifiers close to the data. Similarly to submodels, concept descriptions are not limited in their appearance, therefore the superset $CD^l$ is used.

An instance $aas$ of an AAS is, therefore, defined by the union of the mentioned sets:
\begin{equation}
    aas \in AAS \cup A \cup S^k \cup CD^l 
\end{equation}

The identifiers appear in all sets and are therefore not mentioned separately. They connect the objects of the different sets with each other. However, the nature of identifiers in the AAS data model is mostly the one of foreign keys, which do not link directly to the intended object. 
We define two types of functions on the administration shell. First, a serialization $ser$ transforms each administration shell to a representation in a data format, in particular JSON and XML: 
\begin{math}
ser: AAS \rightarrow D = \{XML, JSON, ...\}
\end{math}
\\Second, a mapping is a transformation $m$ from the data model $AAS$ to the Semantic Asset Administration Shell $SAAS$.
$SAAS$ is defined as
\begin{equation}
    SAAS = AAS_{RDF} \cup A_{RDF} \cup S_{RDF}^k \cup CD_{RDF}^l
\end{equation}

\begin{figure}
        \centering
        \includegraphics[width=\linewidth]{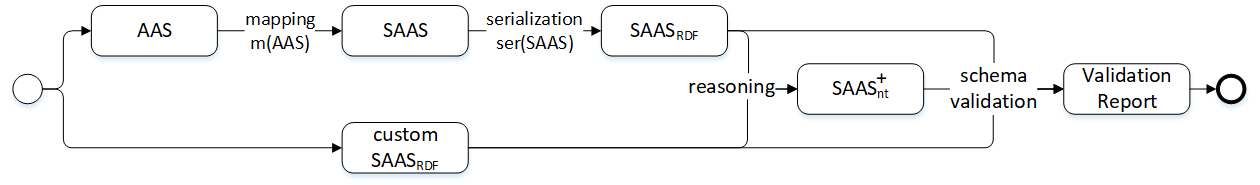}
        \caption{Process steps through the provided modules.} 
        \label{fig:ProcessDiagram}
\end{figure}

Using these definitions, an AAS in XML undergoes several steps (see Fig. \ref{fig:ProcessDiagram}). A created SAAS object using the provided mapping (Section \ref{sec:Mapping}) can be sent to a reasoning engine (Section \ref{sec:Reasoning}) to enrich it with additional facts. Both the native $SAAS_{RDF}$ and the enriched $SAAS_{RDF}^{+}$ can be forwarded to a validation module (Section \ref{sec:Validation}). The validation module creates a validation report, containing the errors and inconsistencies against the SAAS schema. Of course, also otherwise created SAAS objects can be sent to the reasoning or validation modules (bottom lane).

%
%
\section{The SAAS Data Model}
\label{sec:DataModel}

In the following we present the $SAAS$ data model as an RDF ontology\footnote{\url{https://github.com/i40-Tools/RAMIOntology}}. As mentioned, the ontology is an advanced version of the RAMI ontology~\cite{grangel2016towards} and, therefore, the namespace \textit{rami} is used. For each class from~\cite{aasindetail} a corresponding OWL Class has been created and every attribute has been mirrored with either an ObjectProperty or a DataProperty, except for the 'semanticId'. The reason for the later is that 'semanticId' links to the unique identifier for the entity. In RDF, this is the entity URI itself and therefore does not need to be repeated.

\begin{figure}
        \centering
        \includegraphics[width=0.9\linewidth]{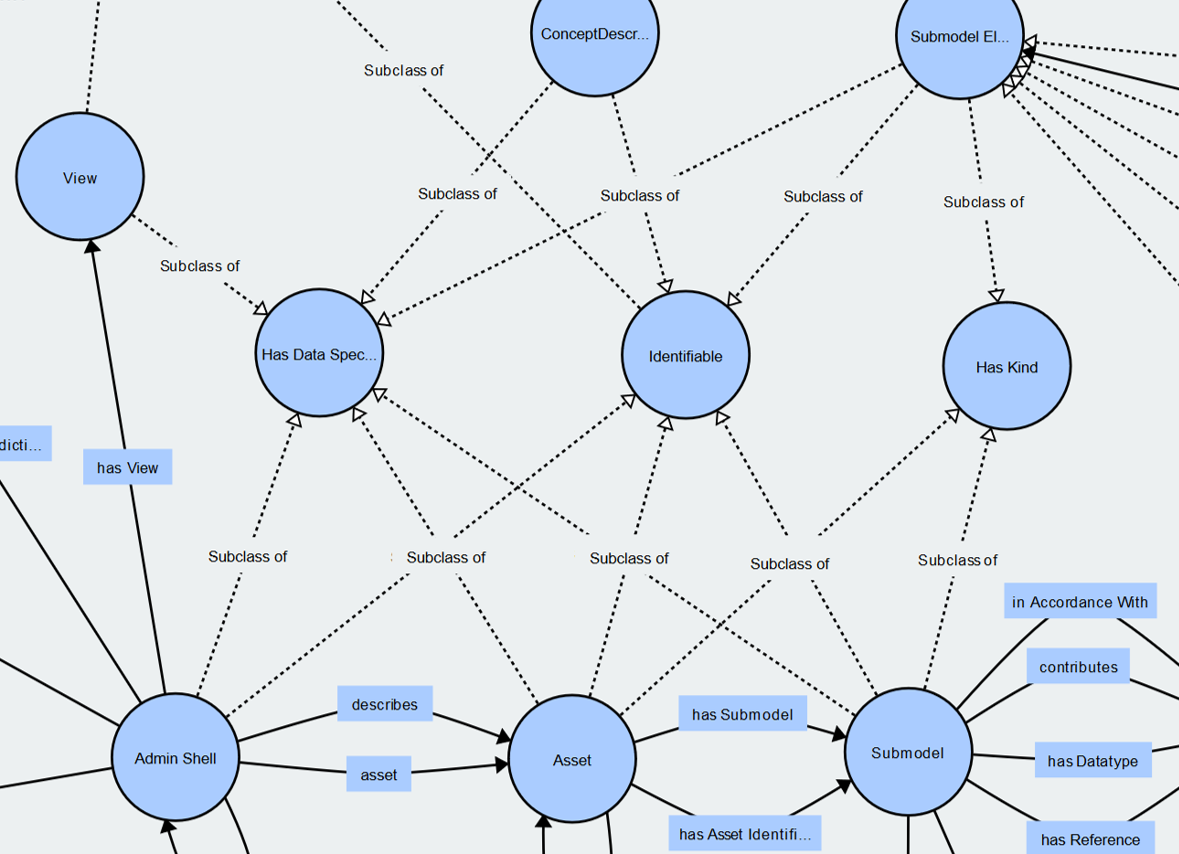}
        \caption{Overview on the most important classes and properties of the SAAS\protect\footnotemark.} 
        \label{fig:SaasRdfDataModel}
\end{figure}

All RDF entities are supplied with (sub)class assertions, labels and comments. The SAAS classes reflect the original ones in most cases and form a subclass hierarchy based on the inheritance specification of the AAS data model. However, neither RDF nor OWL know abstract classes. AAS uses abstract class constructs to partition certain attribute requirements and characteristics. For instance, the 'Has Kind' class covers all realizations, which contain a 'kind' attribute. This attribute encodes whether a certain entity is either referring to a concrete instance (the explicit machine installed in a shop floor) or is related to a whole type (machine type A can be installed in a certain setting). The data model reflects the abstract nature through \textit{:class skos:note ``abstract''} statements.


While the existing schemes for XML and JSON are based on a tree-structure, the RDF data model supports a more generic graph structure. While this might lead to the conclusion that for every model from $AAS_{xml}$ or $AAS_{json}$
a corresponding RDF serialization must be possible, therefore $AAS \subseteq SAAS$, we will show that some limitations exist and actually $AAS \supset SAAS$ is the case.

%
%
\section{Mapping to RDF}
\label{sec:Mapping}

The Administration Shell object ($AAS$) is the root of every Asset Administration Shell. Listing \ref{lis:RaspberryPiXMLexample} shows an example XML snippet. As the root entity, it is also the entrypoint for traversing the SAAS graph. A native mapping is always possible if the identifier is already applied in the form of an URI. However, also International Registration Data Identifiers (IRDI) and any other custom format is allowed.
\footnotetext{For full visualization see \url{http://www.visualdataweb.de/webvowl/\#iri=https://raw.githubusercontent.com/i40-Tools/RAMIOntology/master/rami.ttl}}
 While IRDIs in case of the wide-spread eCl@ss system can -- with significant additional efforts -- being mapped to URIs, this is in general a very hard and error-prone challenge\footnote{For instance, templates for eCl@ss IDs, e.g. 26-04-07-02 (High-voltage current), may map to \url{https://www.eclasscontent.com/index.php?id=26040702}}. This becomes even harder when regarding proprietary or custom identifiers. In addition, custom identifiers may contain special characters as spaces or several hash signs. These characters are percent encoded (\textit{\#} $\rightarrow$ \textit{\%23}, changing the appearance of identifiers. As a result, only native URI identifiers can be mapped without risk, not only for AAS identifiers but also for the other sets in the following. 

A consequence of this decision is also that the 'Has Semantics' class and the 'semanticId' property of the AAS data model becomes native to all objects. Moreover, it implies that all URIs are not only uniquely identifying its data object but also supply the semantic definition of their meaning. This rather strict requirement can be further aligned with the Linked Data Principles if URIs are also enforced to point to actual resources. However, dereferencable URIs are not a requirement for now but should be seen as a preferable best practice. 

The asset objects ($A$) constitute the link from the AAS to the real-world thing. As assets themselves only contain a very brief description, only the class assertions (rdf:type), the name (rdfs:label), descriptions (rdfs:comment) and the kind attribute are translated to $A_{RDF}$. 

Submodels ($S$) and SubmodelElements are the core information carrier of the Asset Administration Shell. The basic structure of the submodel serves as a bracket for several SubmodelElements. Abstract SubmodelElements can be realized by Operations, ReferenceElements, Files, binary objects (Blob) and Properties. Properties have further attributes such as a key, value, value type and several others. In order to align the Property class with the graph model of RDF, each instance is transformed to a respective rdf:Property. Therefore, a distinct class 'Property' does not exist in SAAS. The alternative usage of n-ary relations, which would further allow the linking of more attributes to the relation, was discarded in order to sustain cleaner graphs. Consequently, not all Property objects can be translated to the SAAS model.

\begin{lstlisting}[language=XML, caption=XML serialization of the Raspberry Pi AAS\protect\footnotemark., float, floatplacement=b, label=lis:RaspberryPiXMLexample]
<?xml version="1.0"?>
<aas:aasenv xmlns:IEC61360="http://www.admin-shell.io/...">
  <aas:assetAdministrationShells>
    <aas:assetAdministrationShell>
      <aas:idShort>RaspberryPiModel3B+</aas:idShort>
      <aas:identification idType="URI">
        http://iais.fraunhofer.de/.../raspberry_pi_3b_plus
      </aas:identification>
      <aas:assetRef>
        <aas:keys>
          <aas:key type="Asset" local="true" idType="URI">
          https://iais.fraunhofer.de/.../rspbry/755003377
          </aas:key>
        </aas:keys>
      </aas:assetRef>
      ...
   </aas:assetAdministrationShell>
 </aas:assetAdministrationShells>
</aas:aasenv>
\end{lstlisting}
\footnotetext{Examples can be found at \url{https://github.com/i40-Tools/RAMIOntology/tree/master/AssetAdministrationShell_examples}}

Mainly, attributes and properties are converted to triples and identifiers are restricted to URIs. Therefore, all identifiers of attributes become globally valid, as URIs are globally valid. It has been deliberately decided against n-ary constructs with blank nodes and an explicit property class, which would have been closer to the XML and JSON influenced data model. The reason is that an thereby created graph increases in complexity while its comprehensibility significantly decreases and the information content stays the same.

\begin{lstlisting}[float, caption=Example RML TriplesMap excerpt., label=lis:AssetShallMap]
_:AssetShellMap a rr:TriplesMap ;
	...
	rr:subjectMap 	[
		rml:reference "identification" ;
		rr:class rami:AdminShell ] ;
	rr:predicateObjectMap 	[	
		rr:predicateMap [ rr:constant rdfs:label ] ;
		rr:objectMap 	[	
			rml:reference "idShort" ;
			rr:termType rr:Literal ;
			rr:datatype xsd:string ] 
	] ; ...
\end{lstlisting}

\begin{lstlisting}[float, caption=Equivalent representation to Listing \ref{lis:RaspberryPiXMLexample} as RDF/Turtle.\protect\footnotemark, label=lis:RaspberryPiTurtleExample]
<http://iais.fraunhofer.de/en/aas/examples/raspberry_pi_3b_plus> a rami:AssetShell;
  rdfs:label "RaspberryPiModel3B+";
  rami:hasAsset "http://iais.fraunhofer.de/en/aas/devices/rspbry/755003377"; ...
\end{lstlisting}

Concept description objects ($CD$) serve as local dictionaries for used entities. As the proliferation of definitions and metadata directly with the productive data eases its interpretation, Concept Descriptions increase the degree of interoperability between AAS providing and consuming components. RDF and Linked Data however propagate the usage of dereferencing URIs in order to retrieve metadata. In that sense, Linked Data conventions can reduce the amount of transmitted data. On the other hand, not all relevant Industrie 4.0 components are able to actively request such metadata. The possibility to independently open outgoing interactions beyond the restricted shop floor network is usually also a security risk and is not a good practice. Therefore, Concept Descriptions are a valuable feature to ship metadata and to ensure a common understanding on the shipped AAS.
\footnotetext{Full example: \url{https://github.com/i40-Tools/RAMIOntology/tree/master/rml_mapping/mapping_examples}}
The mapping itself is provided as RML TripleMaps (see Listing \ref{lis:AssetShallMap}) and can be executed with the open-source tool RMLMapper\footnote{accessible at \url{https://github.com/RMLio/rmlmapper-java}}.

%
%
\section{Reasoning}
\label{sec:Reasoning}

RDF and RDFS already contain trivial entailment rule sets\footnote{\url{https://www.w3.org/TR/rdf11-mt/}}. As RDF and RDFS are very general vocabularies, the allowed reasoning focuses on the syntactic position (subject, predicate, object) of entities in RDF graphs. For instance, the information that \textit{p} is an instance of the class Property can be inferred from the fact that a triple with \textit{p} at the predicate position exists. Although rule entailments of this kind are certainly correct, the created amount of explicit data increases significantly while the information content stays nearly the same. 

In order to illustrate the power of reasoning based on the SAAS, selected rule sets using owl:sameAs and rdfs:subClassOf properties have been prepared. The rules are encoded in N3 according to Stadtm{\"u}ller et al. in order to use their Linked Data Integration and Reasoning Engine~\cite{stadtmuller_datafu_2013}. In addition to the two entailment regimes, both consisting of several single rules\footnote{rdfs9 and rdfs11 from \cite{hayes2014rdf}, transitivity, symmetry and replaceability characteristic for owl:sameAs}, the SAAS ontology with its inherent axioms is integrated on the fly. Section \ref{sec:resoning-outcome} presents the results.

%
%
\section{Schema Validation}
\label{sec:Validation}

The AAS presents a closed-world model. As such, the definitions of classes and properties must be regarded as restrictions and simply reusing properties, which were introduced for class A, for class B usually causes a violation of the model. RDF on the other hand does by default allow all not excluded patterns. Nevertheless, industrial use cases require verifiable statements on the data content but also its structure. 

The Shapes Constraint Language (SHACL)~\cite{Knublauch.2017} introduces a W3C recommendation for validation mechanisms on RDF graphs. The definition of required attributes, cardinality of relations or datatype restrictions in the form of shapes is an important aspect to enable data quality assurance in any productive system. Some tools are already created to assist the creation of SHACL shapes, e.g. a Prot\'eg\'e plugin and as a  part of TopBraid Composer. As SHACL shapes are also defined in RDF, they share the same format as the validated data in contrast to e.g. plain SPARQL Rules. This eases the required technology stack and reduces the amount of used libraries. 

The SAAS supplies respective shapes for all its classes\footnote{\url{https://github.com/i40-Tools/RAMIOntology/tree/master/schema}}. These shapes mainly check for mandatory properties but also check the existence of label and comment annotations. In addition, the shapes are essential in order to check the incoming data during the exchange of Asset Administration Shells. Furthermore, the shapes can also be used to describe input and output specifications. For instance, an Industrie 4.0 component can postulate that its API requires data objects conforming to the Asset Shape and will output Submodel objects as defined by the Submodel Shape.

%
%
\section{Use Cases}

We use three different Asset Administration Shells in order to evaluate our approach. All of them are reflecting the specifications from~\cite{aasindetail} and are in the AASX file format. The corresponding descriptions are included in XML files contained in the AASX files.

\textbf{\textit{Raspberry Pi.}}
The first Asset Administration Shell represents a Raspberry Pi 3B+ (see Listing \ref{lis:RaspberryPiXMLexample}). Three Submodels are included, namely one for the technical characteristics, one containing documentation material as the product sheet and a usage manual, as well as one submodel explaining the asset itself. Here, the asset is one specific Raspberry Pi (kind=instance) and not referring to the type of product of all Raspberry Pis, which have been produced or will ever be produced (kind=type). Therefore, the description is only valid for one and only one Raspberry Pi. The AAS delivers 52 SubmodelElements.

\textbf{\textit{Automation Controller.}}
AAS2 describes an electronic controller for automation facilities. As it is not approved as an official artifact, the providing company as well as its details can unfortunately not be published. AAS2 contains one asset, three submodels and more than 100 SubmodelElements.

\textbf{\textit{Multi-protocol Controller.}}
The third use case (AAS3) represents an internet-capable controller unit with multiple protocol support. Like AAS2, this Asset Administration Shell is not officially published yet. However, none of the authors of this paper was involved in the creation of either AAS2 or AAS3. The third use case includes one Asset with eight Submodels and more than 150 SubmodelElements.

%
%
\section{Experimental Evaluation}

We evaluate the AAS to SAAS mapping by examining the results and the performance of the three use cases (see Table~\ref{tab:results}). As a reference to estimate the information coverage, the number of XML nodes of the AAS serializations are provided. In addition, the amount of unique leaves of the three XML trees are noted, as these numbers better reflect the single information content of the AAS. Table~\ref{tab:results} also presents the numbers of generated triples by the RMLMapper. The comparison indicates, as already mentioned, that not the whole expressiveness of AAS can be transported to the SAAS version. This is due to the fact that some constructs can not being represented sufficiently in RDF (for instance the Property class) but also many original entities contain redundant information. Especially the ConceptDescriptions repeat many attributes, which are collapsed by the mapping process and only added once.

\begin{table}[h]
    \centering
\begin{adjustbox}{width=1\textwidth} \begin{tabular}{r|c|c||c|c|c|c|c}
                & \#XML Leaves / & AAS & \#Triples  & SAAS & SAAS & SAAS & SAAS  \\ 
                & \#XML Nodes & (XML) & & (XML) & (nquad) & (turtle) & (JSON-LD) \\ \hline 
                
    RaspberryPi & 1161/2864 & 148 KB & 510 & 40 KB & 86 KB & 32 KB & 51 KB \\ 
    AAS2 &  925/2604 & 91 KB & 459 & 17 KB & 58 KB & 12 KB & 20 KB \\ 
    AAS3 &  2651/6743 & 313 KB & 1154 & 43 KB & 156 KB & 31 KB & 52 KB 
    \end{tabular}
    \end{adjustbox}
    \caption{Results of the SAAS mapping and RDF serialization. 
    }
    \label{tab:results}
\end{table}

\vspace{-2em}
\subsection{Mapping Time}

\begin{figure}[h]
    \centering
    \includegraphics[width=0.7\linewidth]{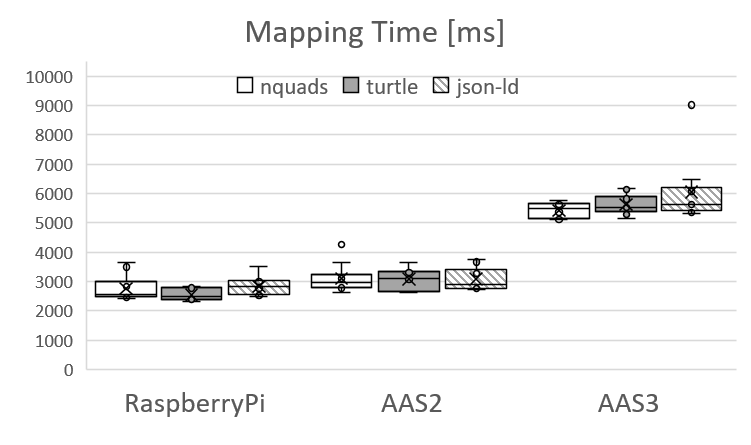}
    \caption{Mapping times for the three Asset Administration Shells.
    }
    \label{fig:MappingTime}
\end{figure}

The necessary overhead in terms of computation time measured in milliseconds is presented in Fig. \ref{fig:MappingTime}, in addition to the average mapping times outlined in the last column of Table \ref{tab:results}. The time was measured on a regular laptop (Win10, 16GB, Intel i5-7300 2,60 GHz) using a bash emulation. The different RDF serializations do influence the execution time, indicating that the writing is not the bottleneck. While the average mapping time of the Raspberry Pi AAS (2,7 sec) and AAS2 (3,1 sec) are rather close, the duration for AAS3 (5,7 sec) is significantly higher. The variation between the selected use cases reflects the differences in their XML file size. This could indicate that the overall behavior is nearly linear. However, each of the 19 TripleMaps leads to a reloading and reiteration of the whole XML file. Overcoming this expensive process would speed up the process significantly but is out of the scope for this paper.

\subsection{Data Overhead}
RDF is in general not an effective data format in terms of storage efficiency. Nevertheless, the syntax requirements of the AAS and especially its XML schema create already significant overhead for the original AAS model. As depicted in Table \ref{tab:results}, all RDF serializations reduce the necessary storage size. Especially noteworthy is the difference between the original XML file size and the RDF/XML serialization. This is mostly due to the usage of namespaces in the RDF/XML version, which reduces the noted URIs. It should be mentioned that for all serializations the mapping step (m) and the serialization (ser) were executed directly by the mapping engine.

Nevertheless, the resulting costs in terms of storage requirements and communication bandwidth do not exceed the ones created by the original Asset Administration Shells. Consequently, all devices and scenarios capable of handling AAS are also sufficient for the operation of SAAS. Furthermore, the possible serialization of SAAS as both XML and JSON should enable AAS implementations to quickly adapt to SAAS objects in their original file format.  

\subsection{Reasoning}
\label{sec:resoning-time}
\label{sec:resoning-outcome}

Three different rule sets have been applied to all use cases. All rule sets contain a web request to the ontology source file in order to load the class hierarchy and any other relevant axioms of the data model itself. The first one also adds several rules reflecting the symmetry and transitivity of owl:sameAs as well as the fact that same instances share all properties and annotations of each other. The second rule set contains subclass statements as encoded by the rules rdfs9 and rdfs11~\cite{hayes2014rdf}. The third set combines both to the most expressive reasoning set. Table \ref{tab:reasoning-results} gives an overview of the amount of created triples. \textit{rdfs:subClassOf}, \textit{owl:sameAs} and the combination of both entailments are shown with the amount of uniquely added triples and the average reasoning time.

\begin{figure}[]
\centering
    \begin{minipage}{.5\textwidth}
        \centering
        \includegraphics[width=1\linewidth]{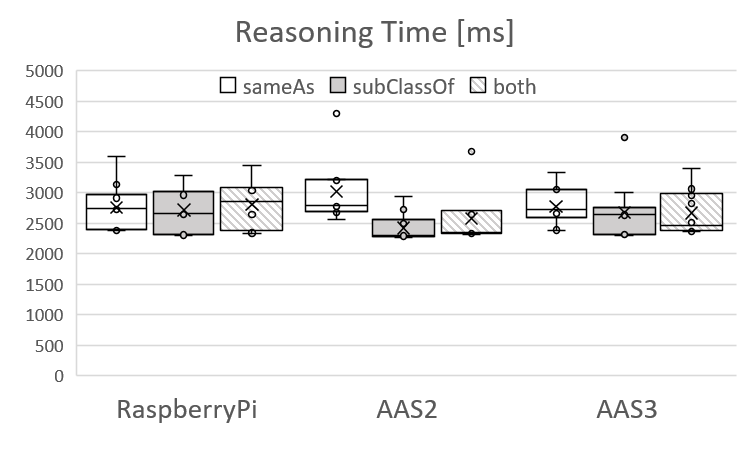}
        \captionof{figure}{
        SAAS Reasoning duration.
        } 
        \label{fig:ReasoningTime}
    \end{minipage}%
    \begin{minipage}{.5\textwidth}
        \centering
        \includegraphics[width=1\linewidth]{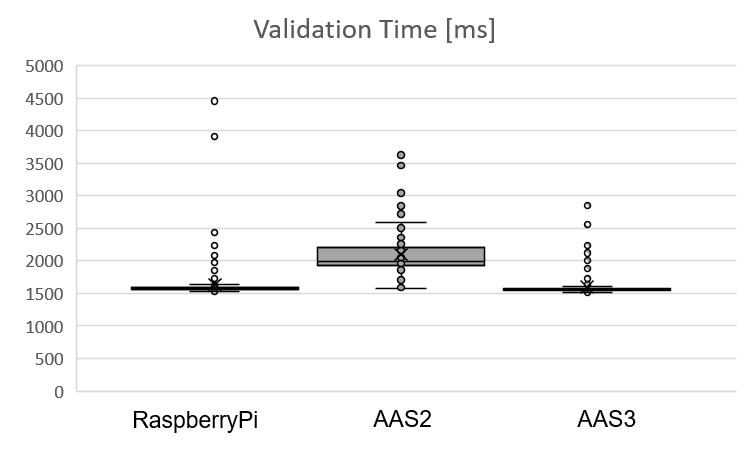}
        \captionof{figure}{Schema validation performance.
        } 
        \label{fig:ValidationTime}
    \end{minipage}
\end{figure}

We use the Linked Data-Fu engine~\cite{stadtmuller_datafu_2013}. The preparation of the reasoning engine, involving the parsing of the rule files, takes around 1 second. The following web request, the download of the ontology, the evaluation of the rules and the serialization to a n-triple file is then executed. The duration distribution of ten repetitions is shown in Fig. \ref{fig:ReasoningTime}. One can see that the whole process takes between 2,3 and 3,3 seconds, nearly independently of the amount of inputs (AAS3 is significantly larger than the graph for the Raspberry Pi) and the expressiveness of the rule sets (the second set is leading to way less results than the others).

As the rule sets are only regarding the structure of the ontology, the inferencing of context-dependent knowledge is not yet possible. In order to reach productively usable information, domain-specific axioms tailored to the actually contained or expected data is necessary. However, we can show that the reasoning process with complex rules is applicable in an acceptable amount of time.

\begin{table}[t]
    \centering
    \begin{adjustbox}{width=1\textwidth}
    \begin{tabular}{r|c|c|c|c|c|c|c}
                &  Triples    & sameAs      & sameAs & subClassOf    & subClassOf & both        & both \\ 
                &  (original)   & (triples) & (time) & (triples) & (time)   & (triples) & (time) \\ \hline
                
    RaspberryPi & 510           & 959         & 2,760 ms & 771       & 2,719 ms & 1,217       & 2,808 ms \\
    AAS2        & 459           & 452         & 3,057 ms & 367       & 2,368 ms & 570         & 2,313 ms \\
    AAS3        & 1154          & 1,115       & 2,776 ms & 818       & 2,677 ms & 1,343       & 2,668 ms
    \end{tabular}
    \end{adjustbox}
    \caption{Added triples by the different rule sets.}
    \label{tab:reasoning-results}
\end{table}

\subsection{Schema Validation}

The evaluation times of the SHACL shapes are shown in Fig. \ref{fig:ValidationTime}. On average, the execution of all shapes takes 46,2 seconds and the execution of one single shape 1,8 seconds. All shapes have been executed a total of ten times.

About 2 seconds are required for setting up the validation tool and parsing the data shape (the Asset Administration Shell) and the single class shape. The size of the Asset Administration Shell has no significant impact on the achieved results. Regarding these conditions, we claim that the necessary effort is acceptable for a typical Industrie 4.0 scenario as the validation itself is not necessary for every restricted devices. This is due to the fact that the validation of data takes either place at development or deployment time where time is not critical. In addition, the validation is important for the higher-level data analytical services which usually run on more powerful machines or are even hosted in the cloud.

%
%
\section{Conclusion and Outlook}
We presented a semantic version of the Administration Admin Shell, a mapping from its XML serialization to any RDF serialization, schema validation shapes and a brief set of reasoning rules. In that sense, we showed the lifting process of the AAS data to a semantic integration layer.

This is one step to an automated integration of Industrie 4.0 components. We showed how existing, non-customized tools can work with the RDF model of the AAS and execute their task without prior configuration. This enables the implementation of real interoperable pipelines and data-driven workflows, not only on the data format and syntax level but also regarding the meaning of the data. Furthermore, the examined overhead of the SAAS model and showed that the requirements do not exceed the requirements set by the original AAS model.

The mapping provided in this paper outlines the data lifting to the SAAS RDF model. The lowering of RDF to the original AAS data model has not yet been achieved. Furthermore, the main benefit of the semantic model is, besides its formalized meaning, the interlinking with other definitions and the integration of additional sources. 

For now, only the data provisioning capabilities of the AAS are defined. In the next step, the provisioning and invocation of operations through Asset Administration Shells will be specified.  Using semantically defined descriptions of the respective interfaces, their input and output parameters and the provided services will allow the Industrie 4.0 community to rely on the huge amount of expertise and experience with Web Services and Semantic Web Services in particular~\cite{pedrinaci2011adaptive,pedrinaci2012semantic}. This way, the goal of truly interoperable and flexible manufacturing workflows, where software and hardware, materials and products, costumers and suppliers form on demand information chains, benefits from the huge amount of existing research in the area.

We will further extend our work in order to keep the semantic models aligned with the progress of the Asset Shell specification. Furthermore, we provide feedback and outline established best practices to the manufacturing community. Furthermore, we see two main challenges which must be tackled by the semantic community. First, the core potential of the semantic web -- the seamless integration of heterogeneous devices, services and data sources -- still lacks sufficient numbers of implemented use cases and deployed scenarios in practice. Second, the reoccurring discussion on identifiers in distributed settings is a huge chance for the established practices of the Semantic Web and Linked Data in particular. However, the benefits of (dereferencable) URIs are still underestimated in the manufacturing community, mostly because of missing experiences.

%
%
%
\bibliographystyle{splncs04}
\bibliography{references}
\end{document}